\documentclass[sensors,article,accept,pdftex,moreauthors]{Definitions/mdpi} 
\firstpage{1} 
\pubvolume{26}
\issuenum{4}
\articlenumber{1127}
\pubyear{2026}
\copyrightyear{2026}
\externaleditor{Basheer Qolomany, Jacques Bou Abdo and Liaquat Hossain} % More than 1 editor, please add `` and '' before the last editor name
\datereceived{25 December 2025} 
\daterevised{26 January 2026} % Comment out if no revised date
\dateaccepted{5 February 2026} 
\datepublished{ } 
\Title{Assessing Low Back Movement with Motion Tape Sensor Data Through Deep Learning}

\Author{Jared Levy $^{1}$\orcidJ{}, Aarti Lalwani $^{1}$\orcidL{}, Elijah Wyckoff $^{2}$\orcidE{}, Kenneth J. Loh $^{2}$\orcidA{}, Sara P. Gombatto $^{3}$\orcidB{}, Rose Yu $^{1}$\orcidR{} \linebreak  and Emilia Farcas $^{4,}$*\orcidC{}} 
\AuthorNames{Jared Levy, Aarti Lalwani, Elijah Wyckoff, Kenneth J. Loh, Sara P. Gombatto, Rose Yu, and Emilia Farcas}

\address{%
$^{1}$ \quad Computer Science and Engineering, University of California San Diego, La Jolla, CA 92093, USA; jalevy@ucsd.edu (J.L.); adlalwani@ucsd.edu (A.L.); roseyu@ucsd.edu (R.Y.)\\
$^{2}$ \quad Active, Responsive, Multifunctional, and Ordered-Materials Research (ARMOR) Laboratory, Department of Structural Engineering, University of California San Diego, La Jolla, CA 92093, USA; ewyckoff@ucsd.edu~(E.W.); kenloh@ucsd.edu (K.J.L.)\\
$^{3}$ \quad School of Physical Therapy, San Diego State University, San Diego, CA 92182, USA; sgombatto@sdsu.edu (S.P.G.)\\   
$^{4}$ \quad Qualcomm Institute, University of California San Diego, La Jolla, CA 92093, USA}

\corres{Correspondence: efarcas@ucsd.edu (E.F.)}

\abstract{Back  %MDPI: Please define all abbreviations when they first appear in the abstract, main text, and figure or table captions.
pain is a pervasive issue affecting a significant portion of the population, often worsened by certain movements of the lower back. Assessing these movements is important for helping clinicians prescribe appropriate physical therapy. However, it can be difficult to monitor patients' movements remotely outside the clinic. High-fidelity data from motion capture sensors can be used to classify different movements, but these sensors are costly and impractical for use in free-living environments. Motion Tape (MT), a new fabric-based wearable sensor, addresses these issues by being low cost and portable. Despite these advantages, novelty and variability in sensor stability make the MT dataset small scale and inherent to noise. In this work, we propose the Motion-Tape Augmentation Inference Model (MT-AIM), a deep learning classification pipeline trained on MT data. In order to address the challenges of limited sample size and noise present within the MT dataset, MT-AIM leverages conditional generative models to generate synthetic MT data of a desired movement, as well as predicting joint kinematics as additional features. This combination of synthetic data generation and feature augmentation enables MT-AIM to achieve state-of-the-art accuracy in classifying lower back movements, bridging the gap between physiological sensing and movement analysis.}

\keyword{deep learning; motion sensing; wearable sensors; back pain; movement classification; generative models; time-series analysis}

\begin{document}

%%%%%%%%%%%%%%%%%%%%%%%%%%%%%%%%%%%%%%%%%%
\section{Introduction}

Back pain %MDPI: Please carefully check and revise the following: 
%(1) The companies/manufacturers of chemicals & reagents, devices, instruments,  
%commercial cell lines/samples/ materials should be indicated together  
%with their city (states abbreviation is required for USA and Canada)  
%and country in their first appearance. Please review the full text carefully and add the missing information necessarily. 
%Example: oxalic acid (Sinopharm Chemical Reagent Co. Ltd., Shanghai, China).  
% (2) For any software used in the study, please provide the version number. If the version number is not available, please provide the weblink and access date, instead.
% Authors' comments: added the required info in sect 3 Materials.
 is a widespread and debilitating health condition that affects individuals throughout the world {\cite{chen-lbp}}. According to the World Health Organization, back pain is one of the top five leading causes of disability worldwide \cite{paho2025}. Around 70--85\% of adults experience low back pain (LBP) at some point in their lifetime \cite{andersson1999}; among them, an average of 62\% still experience pain one year after initial onset and 60\% will suffer from a recurrent episode of pain \cite{hestbaek2003}. The total costs of LBP in the US exceed \$100 billion per year, including lost wages due to inability to work \cite{katz2006}.

Physical therapy is effective for {preventing and treating} LBP {\cite{Saragiotto16, Steffens-prevention}}, but adherence to prescribed exercises is low \cite{anar2016}. Barriers include the lack of personalization of exercises to a given patient's needs, {reduced engagement due to the gap the patients feel between supervised sessions in clinic and performing exercises at home without support or oversight~\cite{Palazzo2016-rs, Boutevillain2017-cf}}, and the difficulty in determining whether patients are correctly adhering to the physical therapists' recommendations at home{---the clinician does not know what the patients are doing at home, but also LBP patients do not know whether they are performing exercises correctly due to impaired proprioception \cite{Zheng2022-tl,Ghamkhar2019-ze, Poesl2023-qy}}. 

Traditionally, determining the severity of the LBP and specific movements that exacerbate the LBP, prescribing the recommended exercises, and monitoring the progress throughout physical therapy are all done through clinical evaluations and interventions {to enhance strength, stability, and mobility and ultimately mitigate pain and disability~\cite{HENCHOZ2008533, Nelson1995-sv, Marich2018-ah}}. These evaluations are typically conducted in person, and can be time consuming and difficult to conduct remotely. With the increasing need for remote healthcare solutions, accurately monitoring and assessing movement patterns at home has become essential for personalized evaluation, development of treatment plans, and prescription and monitoring of therapeutic interventions. {Remote monitoring that quantifies the diverse activities patients engage in free-living environments can help identify movement patterns linked to LBP \cite{Silva2021-yh} and monitor adherence to the prescribed home exercises.}

Advances in sensor and deep learning technologies have provided promising alternatives for the remote monitoring and analysis of movements {\cite{Alschuler2011-gr, Schaller2016-zn, Dutta2018-ka}}. Prior research has leveraged motion capture (MoCap) \cite{zhu2023stmt}, surface electromyography (EMG) \cite{aarotale2024ml}, and inertial measurement unit (IMU) sensors \cite{garcia2023inertial} to train deep learning models to classify human movement. However, while these models have high-quality data streams, the sensors are expensive and not easily portable for home use and monitoring, making them not accessible to the general population.

\subsection{Problem Setting and Challenges}
% Citation removed for anonymity, (\cite{Spiegel2024MotionTS}, \cite{Lee2024MotionTS})
There is a need for low-cost, portable sensors and Artificial Intelligence (AI) tools to capture low back movements in free-living environments. Motion Tape (MT) has emerged as a promising alternative to other sensors \cite{Lee2024MotionTS}. {MT is a flexible, wearable strain sensor fabricated by depositing a conductive nanomaterial network onto a commercial kinesiology tape (k-tape) substrate. The sensor adheres directly to the skin, allowing it to translate the skin strain induced by joint motion and muscle activation into measurable electrical signals. The sensing mechanism of MT is piezoresistive, where deformation of the nanomaterial network produces changes in electrical resistance that are proportional to the applied strain. During use, time-varying resistance signals are continuously recorded and interpreted as skin deformation associated with movement and muscle engagement \cite{Lee2024MotionTS, Spiegel2024MotionTS}. Because MT is lightweight (on the order of grams), it minimally interferes with natural motion and is suited for prolonged wear outside laboratory settings. MT measurements are inherently influenced by factors such as sensor placement and skin adherence.}
%MT is a wearable sensor that adheres to the skin and measures skin stretch as the underlying bony segments move and muscles engage. During use, changes in electrical resistance within the sensors are continuously recorded to provide an accessible measure of movement and muscle engagement \cite{Lee2024MotionTS, Spiegel2024MotionTS}.

The task of classifying low back movements is challenging due to the multi-segmental nature of the lumbar spine, involving all three planes of motion. Furthermore, noise is inherent in the MT data; the sensors are limited in quantity, as they are custom fabricated in the research lab for this study; and human subject testing with multiple devices and under the supervision of a physical therapist has complex logistics. Thus, obtaining large datasets for AI is not feasible yet. However, training on small datasets leads to overfitting, while noise can cause high bias in model predictions \cite{varoquaux2017cross}.

\subsection{Our Approach}
In this paper, we aim to achieve highly accurate movement classification using MT data. Given a dataset of MT data streams from multiple sensors placed on the low back, the primary purpose of this study is to train a model to distinguish between one of six low back movements, namely standing extension, forward flexion, left/right lateral bending, and seated left/right rotation. Our secondary purpose is to create a machine learning (ML) classification pipeline that addresses the problems that arise from a noisy and limited dataset. This is the first study of deep learning from MT data for low back movements. 

\begin{figure}[H]
  % \centering 
  \includegraphics[width=\textwidth]{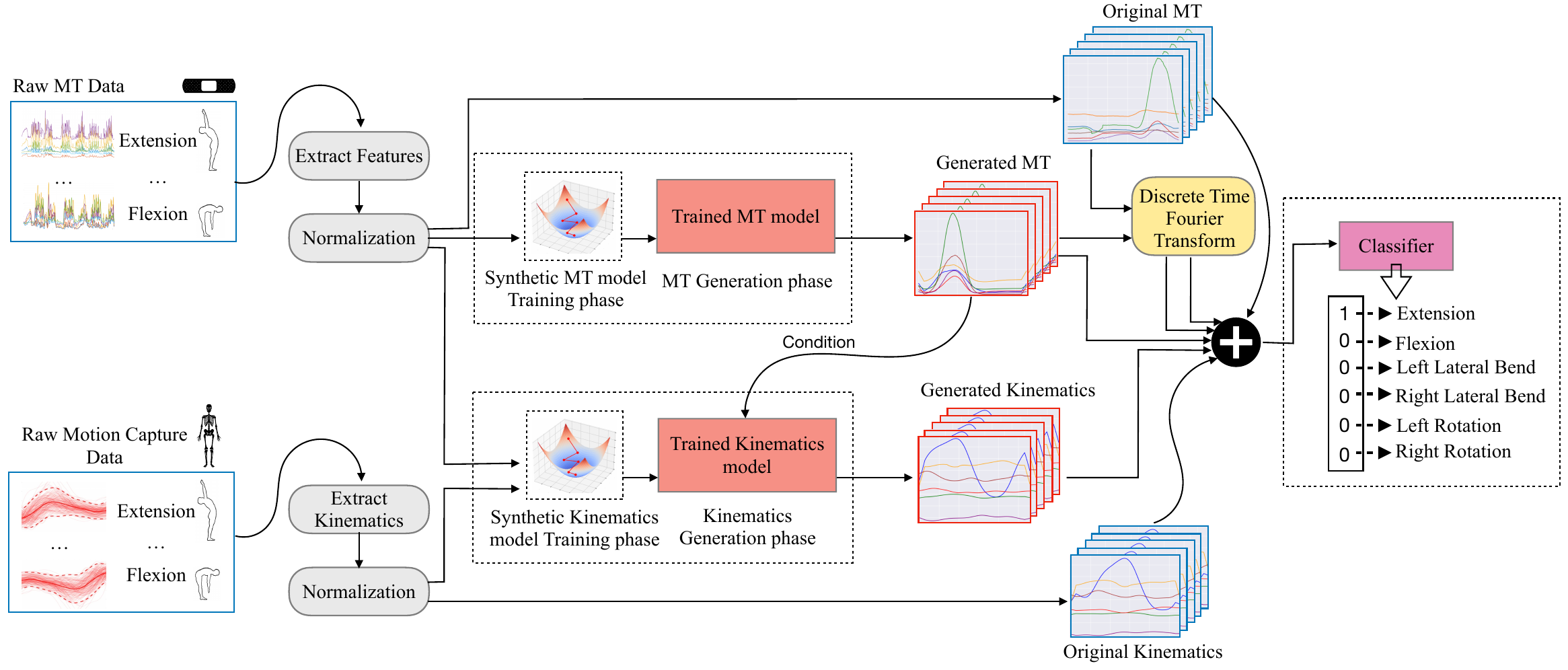} 
  \caption{{Generative and classification data flow for MT-AIM.} Processed MT and MoCap kinematics data are used to train synthetic MT and kinematics generators. The trained models are sampled to create synthetic MT and kinematics data. A DTFT is taken of all MT data. Finally, real, transformed, and synthetic data is all combined before being used to train a classifier.} %MDPI: Note that changes to the position of figures and tables may occur during the layout stage, since all figures/tables should be placed below the first citation, or a few paragraphs lower in order to reduce blanks in the page. 
  % Authors' comments: confirmed
  \label{fig:MT-AIM_pipeline} 
  % \vspace{-3mm}
\end{figure}

We introduce the Motion-Tape Augmentation Inference Model (MT-AIM) %MDPI: Please check if the special font is necessary to be kept. If so, please keep consistent format throughout the whole manuscirpt. If not, please remove the special font.
% Authors' comments: special font has been removed
 a deep learning classification pipeline trained on MT data and leveraging both feature and data augmentation (see Figure \ref{fig:MT-AIM_pipeline}). To tackle the problem of noise, MT-AIM calculates the Discrete-Time Fourier Transform (DTFT) of the MT data, as well as leveraging a generative model to predict respective kinematic angles. In particular, we use marker-based MoCap data taken concurrently with MT data to train a model to generate kinematic angles based on the MT data. These features are then concatenated with the data before classification. Furthermore, an additional generative model is trained to learn the underlying distribution of the MT data and generate synthetic samples conditioned on movement type, thereby increasing the effective size of the training dataset.
\vspace{-6pt}

\section{Related Work}

We consider related work from both ML and clinical perspectives. From a ML classification perspective, dealing with noisy datasets is a known challenge and several pipelines have been proposed to address it. Classical approaches often rely on preprocessing techniques such as data cleaning, feature selection, and dimensionality reduction to mitigate noise and enhance model performance {\cite{pau2023evaluation, goel2019cleanml}}. In cases with many features, methods like Principal Component Analysis (PCA) and Independent Component Analysis (ICA) have been used to reduce the impact of noise by identifying and isolating the most informative features \cite{grunauer2015using}.

In small-scale settings, where the available data may be insufficient for training complex models, techniques such as transfer learning and data augmentation have been applied to improve generalization by leveraging pretrained models or generating synthetic data to expand the training set \cite{chatterjee2022enhancement}. Recent advancements have focused on deep generative architectures that are specifically designed to handle noisy time-series data, including Conditional Variational Autoencoders (C-VAE) \cite{he2024distributional}, Fourier-based diffusion models (Diffusion-TS) \cite{yuan2024diffusionts}, and Generative Adversarial Networks (GAN) such as TimeGAN \cite{yoon2019time}. Furthermore, Bayesian approaches, which model uncertainty in the data and the model parameters, have been shown to improve performance when data is limited or noisy \cite{van2014learning}. In addition, techniques like few-shot learning and active learning have gained attention for addressing small datasets by optimizing the use of available data and actively selecting the most informative samples for training \cite{abdali2022active}.

Many ML solutions exist for synthetic time-series generation, which we can apply to MT data. However, limited research exists related to conditioning the generation upon another time-series, which we apply to kinematic translation. The most similar work is TimeWeaver \cite{narasimhan2024time}. In the paper, Narasimhan et al. use metadata to condition a diffusion model for data imputation. Our problem setting differs in that we are modeling physical relationships over time, rather than correlations with contextual metadata-based features, which motivates our different conditioning strategy.

From a clinical context, marker-based optical MoCap is the prevailing benchmark for movement research, but it demands significant costs and specialized knowledge \cite{Das2023ComparisonOM}. As a result, researchers have explored alternative methods leveraging various sensor data. Classical ML models like support vector machines and gradient-boosted decision trees have been applied to classify arm, head, and neck movements from video and marker-tracking sensors {\cite{Asghar2024ClassificationAM, Jiang2021HeadMC}}. More sophisticated architectures, such as Convolutional Neural Network Long Short-Term Memory (CNN-LSTM) models and Transformers, have demonstrated success in classifying fidgety infant movements, squat variations, and foot movements {\cite{Kulvicius2024DeepLE, Lee2020AutomaticCO, AydinFandakli2024DeepL}}. In domains more closely related to our focus on lower back kinematics, deep learning has enabled the mapping of markerless RGB-D camera data of flexion and extension to specific body segments for spine motion analysis \cite{Wenghofer2024DynamicAO}.

Previous work with MT has also been conducted. In the works of Loh et al. and \mbox{Huang et al.}, strain sensor data collected with MT is used as input to convolutional autoencoders for classifying incorrect movements in warfighter marksmanship exercises and golf swing patterns \cite{loh2022enhancing, huang2023measuring}. Additionally, a three-layer feedforward neural network is constructed to identify subject-specific muscle parameters and predict motion for elbow flexion-extension motion \cite{taneja2022feature}. However, our low back use case presents new challenges for ML due to the multi-segmental nature of the lumbar spine, multi-planar movements that occur during activities, and substantial variability in skin stretch across different planes of motion. This is the first study that uses ML to classify low back movement based on \linebreak  MT data.

As our research shares similar facets to all aforementioned work, we draw upon the most similar and successful to compare and integrate for the MT-AIM pipeline. In particular, we implement C-VAE, Diffusion-TS, and TimeGAN generative models, and XGBoost, Transformer, and CNN-LSTM classification heads.

\section{Materials}

\subsection{Dataset Description}

A total of 10 healthy participants without LBP were recruited to serve as an initial control group to ensure the full range of sensors could be tested and for comparisons in future studies including patients with LBP. Inclusion criteria required participants to be between 18 and 65 years old, able to follow instructions in English, free of LBP in the preceding year, physically capable of walking and bending their backs, and willing to wear standardized testing attire \cite{Lee2024MotionTS}. {The mean age was $22.4 \pm 2.1$, $50\%$ male and $50\%$ female; the detailed demographics and anthropometric measurements are presented in \cite{Lee2024MotionTS}.}

MT sensors were fabricated using a previously established procedure adapted from \cite{lin-MT-21}. Regions of k-tape (Rock Tape, Durham, NC, USA) were masked to define the 7.5 × 40 mm$^2$ %MDPI: We revised the multiplication sign and superscript of the unit. Please confirm the suggested change. % Authors' comments: confirmed.
 sensing area. Graphene nanosheet (GNS) ink was prepared by dispersing graphene within an ethyl cellulose-ethanol solution (solvents from Sigma-Aldrich, St. Louis, MO, USA) deposited onto the k-tape by spray coating. Following the deposition of three spray-coated layers, the resulting GNS sensing element exhibited an initial unstrained resistance of 10 kOhm. Conductive silver paste (Conductor 3 Silver Ink from Voltera, Kitchener, ON, Canada) was applied to each end of the sensing region to form electrodes, then multistrand wires (from Digi-Key, Thief River Falls, MN, USA) were soldered to the electrodes.

\begin{figure}[H]
  \includegraphics[width=0.85\textwidth]{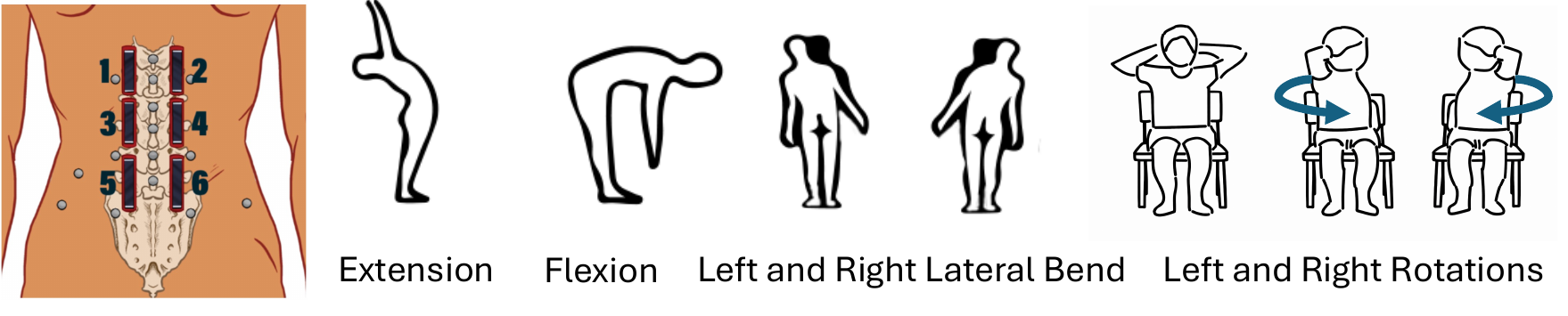}
  \caption{{\textbf{Left:} The placement of six MTs (black stripes) and retroreflective markers used for 3D optical MoCap (gray circles). \textbf{Right:} The six movements tested with human subjects: standing extension, standing forward flexion, standing lateral flexion (left and right), and seated rotation (left and right).}}
  \label{fig:movements} 
  % \vspace{-3mm}
\end{figure}

Six MTs were arranged in a 3 × 2 matrix lateral to the spine, positioned to align with spinal junctions from T12 to S1 (see Figure~\ref{fig:movements}). Retroreflective marker-based 3D optical MoCap provided the reference kinematic data. An experienced physical therapist identified key lumbar spine and pelvic landmarks for marker and MT placement for each participant.  {First, the middle MTs 3 and 4 were placed to cross L2-to-L3 and L3-to-L4 junctions. Above them, the MTs 1 and 2 were placed to cross the T12-to-L1 and L1-to-L2 junctions. Lastly, the bottom MTs 5 and 6 were placed to cross the L4-to-L5 and L5-to-S1 junctions for most participants.} MoCap reflective markers were placed on the spinous processes from T12 to L5 and also to the left and right of L1 and L4 to construct a modified version of a validated multi-segmental spine model, defining upper and lower lumbar segments. Additional markers were placed bilaterally on pelvic landmarks to define the pelvis segment \cite{Lee2024MotionTS}.

The movements tested were standing extension, forward flexion, lateral flexion (right and left), and seated rotation (right and left), as seen in Figure \ref{fig:movements}. Each participant performed three repetitions per each of the six movements, totaling 180 time-series samples around 6 s each. Measurements across all sensors were recorded approximately every 20 milliseconds.
\vspace{-14pt}

\begin{figure}[H]
%  \captionsetup{justification=centering}
%  \centering 
  \includegraphics[width=0.88\textwidth]{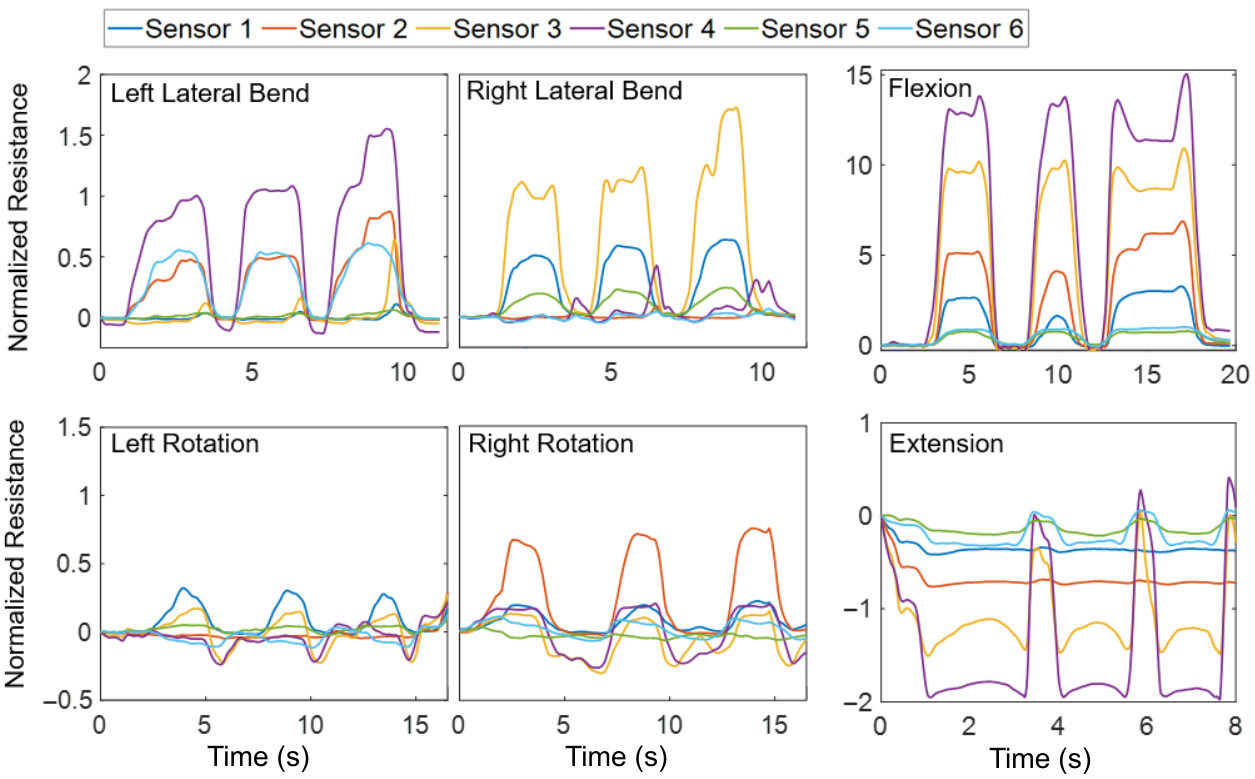}
  \caption{{Example MT normalized resistance for each movement type, repeated three times.}} 
  %MDPI: Figures 3 and 7:
%1. The minus sign of the negative number should be minus (U+2212; “−”) not hyphen (-), we revised it. Please confirm.
%2. The unit second should be “s”, please confirm if it can be revised.
% Authors' comments: we confirm the change to use a minus rather than hyphen; figures 3, 4 and 7 have been changed to use (s) rather than (sec)
  \label{fig:mt-exercises} 
  % \vspace{-3mm}
\end{figure}

{Figure~\ref{fig:mt-exercises} illustrates an example of MT strains measured using all six sensors and three repetitions for each movement type by the same participant. Resistance $R_n$ is normalized to the base MT resistance $R_0$ in an unstretched position before the test: $R_n=(R-R_0)/R_0$, where $R$ is the electrical resistance measured at a point in time.}
\vspace{-6pt}

\subsection{Data Processing}

Motion capture data was processed using Qualisys Track Manager (version 2024.3, Qualisys North America, Inc) to label marker trajectories and interpolate missing data, then transferred to Visual3D (C-Motion, Inc). A multi-segmental spine model was used to calculate lumbar spine kinematic angles using Euler angles between upper lumbar, lower lumbar, and pelvis segments. The resulting angles were exported to MATLAB (version R2023a, MathWorks) for comparison with MT resistance data.  %MDPI: Please state the version number of the software throughout the whole manuscript.
% Authors' comments: added the version and company.

MT resistance data were read into MATLAB R2023a, converted to decimal, and each trial's change in resistance was divided by its baseline. The resulting time series were filtered using a Hampel algorithm to discard outliers \cite{hampel1974influence}; the Hampel filtering used local windows within each sample. Trials with excessive noise, defined as resistance values greater than 10 standard deviations above the mean, were excluded, which resulted in removing less than 2\% of the data. 

Resistance and kinematic data were temporally aligned using synchronized start times, trimmed to match in duration, and normalized between $-1$ and 1 \cite{Lee2024MotionTS}. {The normalization was applied for each movement trial (a trial had three repetitions of the same movement) using that subject's own min/max rather than global statistics. For the strain data, $-1$ corresponds to peak sensor compression and 1 corresponds to peak sensor tension. For kinematics data, $-1$ and 1 correspond to peak movement in each direction.}

\begin{figure}[H]
  \includegraphics[width=\textwidth]{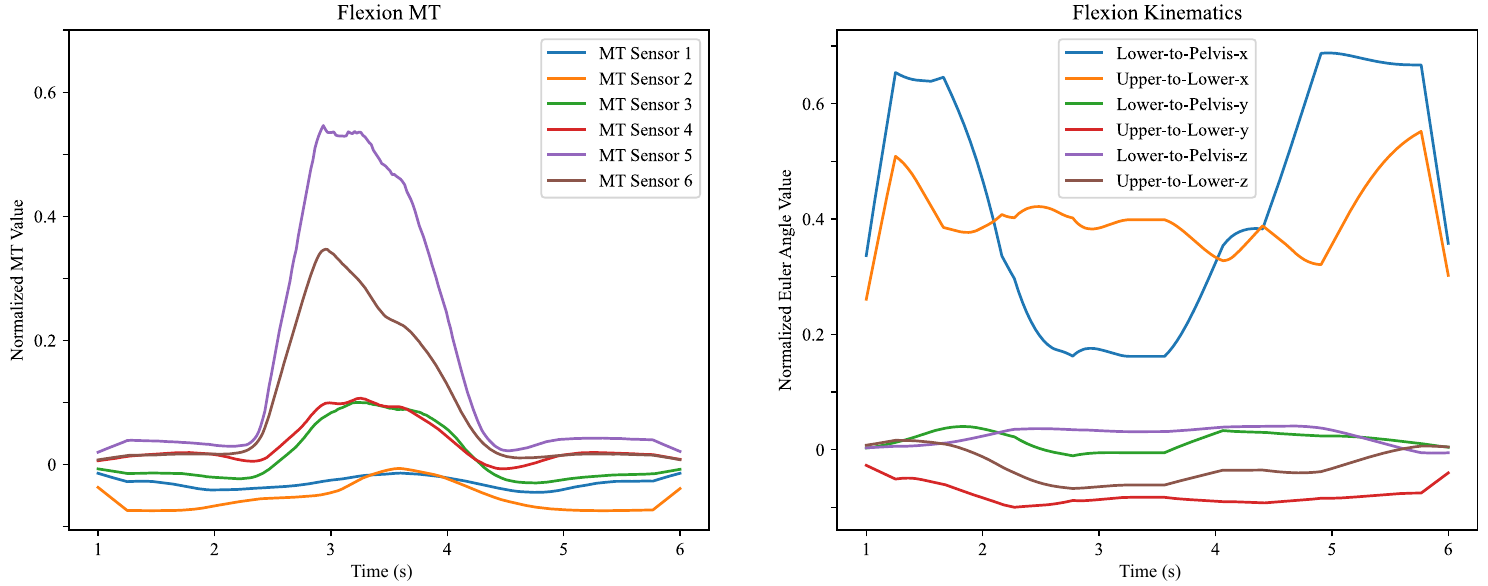} 
  %\caption{\textbf{Left:} Diagram with the placement of six MTs (black stripes) and retroreflective markers used for 3D optical MoCap (gray circles). \textbf{Middle:} Sample of extension MT data showing individual sensors. \textbf{Right:} Sample of corresponding MoCap processed kinematics data showing Euler angles.}
  \caption{{\textbf{Left:} Sample of processed MT data showing individual sensors. \textbf{Right:} Sample of corresponding MoCap processed kinematics data showing Euler angles between the lower lumbar segment (L4--L5) and pelvis segments, and between the upper lumbar segment (L1--L3) and lower lumbar segment (L4--L5).}} %MDPI: The unit second should be “s”, please confirm if it can be revised.
  % Authors' comments: revised to use "s" rather than "sec"
  \label{fig:MT-Mocap} 
\end{figure}

{Figure~\ref{fig:MT-Mocap} shows an example of processed MT and MoCap kinematics data for one flexion movement. MT sensors show tension during flexion, with the highest values for the lower MT sensors 5 and 6, then 3 and 4. This parallels a larger decrease in lower-to-pelvis and upper-to-lower Euler angles for the x axis, which corresponds to sagittal plane flexion.} %(negative angle).}
%The normalization was applied per sample using that sample's own min/max rather than global statistics.
\vspace{-5pt}

\section{Methods: MT-AIM}
% Prompt: Tell us your techniques!  If your paper is develops a novel machine
% learning method or extension, then be sure to give the technical
% details---as you would for a machine learning publication---here and,
% as needed, in appendices.  If your paper is developing new methods
% and/or theory, this section might be several pages. If you are combining existing methods, feel free to cite other packages and papers and tell us how you put them together; that said, the work should stand alone for someone in that general machine
% learning area. \emph{Lack of technical details, such that the soundness of the methods can be verified, is a major reason that otherwise strong-looking papers are scored low/rejected.}

We introduce a deep learning framework designed to assess low back movement with MT sensor data. To address the challenges posed by limited sample size and signal noise within the MT dataset, 
%
% Our framework is structured into three core components — data augmentation, feature augmentation, and classification — each contributing to improved model generalization and robustness. A high-level overview of the pipeline is presented in Figure (\ref{fig:3}), with dotted regions highlighting training and sampling phases of our augmentation modules.
we leverage conditional generative models to augment both the size and feature set of the data. This leads us to our solution, MT-AIM, which combines these data and feature augmentation techniques to optimally train our classification model. Specifically, the data augmentation model generates synthetic MT samples, which are then added to the original training set. Next, we apply DTFT transformations and generate kinematic joint-angle predictions conditioned on the MT samples, which are then concatenated with their corresponding original samples. The resulting augmented dataset is then used to train the classification models. See Figure \ref{fig:MT-AIM_pipeline} for an overview of our approach. For all the models, hyperparameters were selected via an exhaustive grid search.

\subsection{Generative Modeling for Data Augmentation}
Classification models trained on datasets with limited sample sizes and noisy feature representations often suffer from overfitting and poor generalization to unseen data. A common and well-supported approach to mitigate this issue is the incorporation of synthetic data during training. To generate additional synthetic MT samples, we implemented and compared three generative models from related works: C-VAE \cite{pagnoni2018conditional}, Diffusion-TS \cite{yuan2024diffusionts}, and TimeGAN \cite{yoon2019time}. Each model was conditioned on the type of movement to produce class-relevant MT time-series samples. 

\begin{itemize}
        \item For the C-VAE model, the search space included the number of decoder layers and the dimensionality of the hidden state. The optimal configuration consisted of 4 decoder layers and a hidden state dimension of 12.
        \item For the Diffusion-TS model, grid search was performed over the number of sampling steps, hidden state dimensionality, and the number of decoder layers. Given the model’s complexity, the best performance was achieved with 500 sampling steps, a hidden state size of 64, and 12 decoder layers.
        \item For the TimeGAN model, we optimized the number of additional training iterations for the generator module, the latent space dimensionality, and the number of embedding layers. While model performance was relatively stable across parameter settings, we report the results using a 2-to-1 generator--discriminator iteration advantage, a latent dimension of 12, and 3 embedding layers.
\end{itemize}

We used the best performing movement-conditioned generative model to generate 20 synthetic samples per movement type, resulting in 120 synthetic MT samples in total, which were subsequently used to augment the training dataset for classification.

\subsection{Generative Modeling for Feature Augmentation}
To further improve the representational capacity of the model and enhance classification performance, we expanded the feature set associated with each sample. Richer feature representations are particularly important when working with noisy and low-dimensional physiological data, where subtle movement patterns may be difficult to capture using raw signals alone. Prior to augmentation, the original features consisted of six post-processed MT strain time series, one per placed MT (see Figure~\ref{fig:movements}), obtained by normalizing baseline adjusted resistance signals and filtering outliers. As an initial step, we then applied DTFT to each signal, producing 6 additional time-independent frequency-domain features, making it 12 features in total.

To further enrich the feature space with bio-mechanically meaningful information, we modified the previously used conditional generative models---C-VAE, Diffusion-TS, and TimeGAN---to predict joint-angle kinematics of the upper lumbar, lower lumbar, and pelvis segments, conditioned on MT data. Instead of conditioning on a scalar movement type, we need to condition the generative model on the entire time series of MT sensor data.  Essentially, the model acts as a ``translator'' to map the limited MT data to the corresponding joint-angle kinematics. 

\begin{figure}[H]
  % \centering 
  \includegraphics[width=\textwidth]{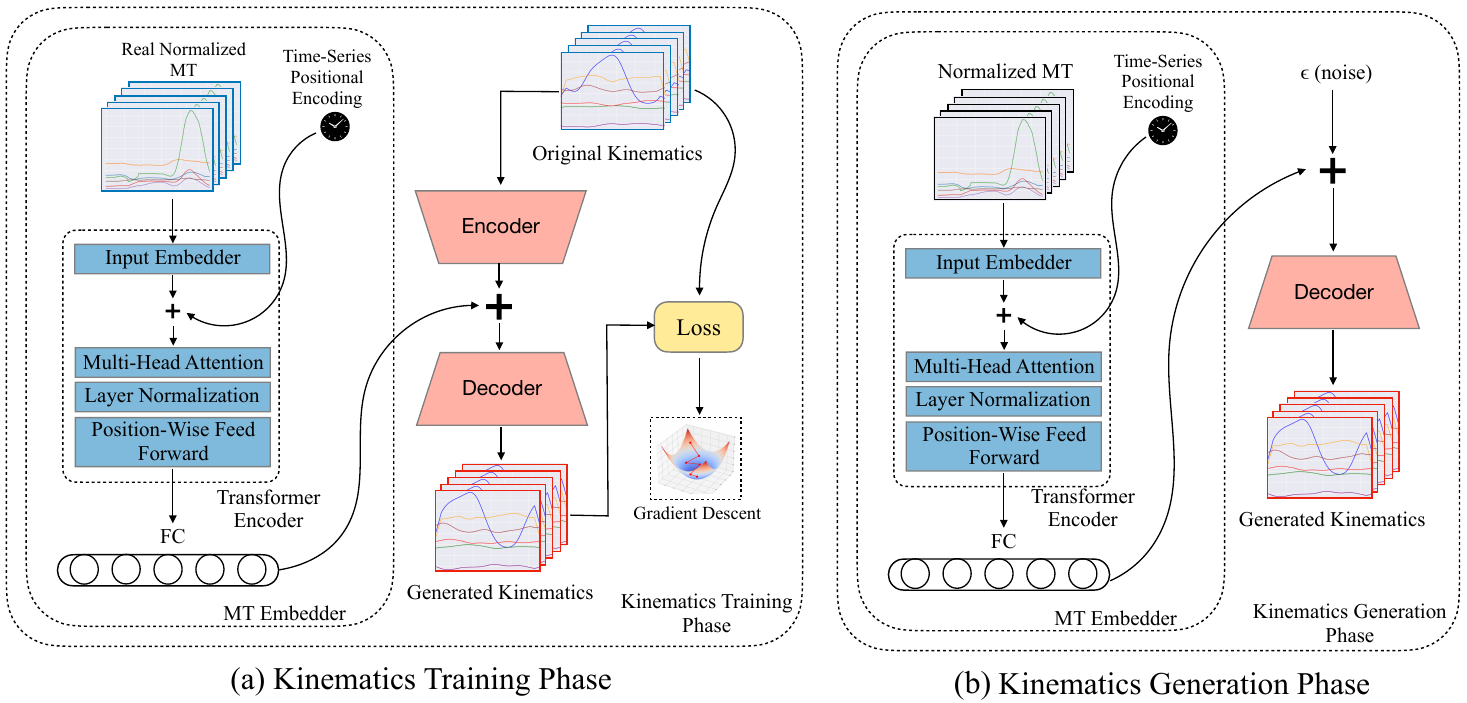} 
  \caption{{General data flow and architecture for the kinematics generators.}
  {(\textbf{a}) An MT-conditioned Transformer Encoder produces an embedding that is summed with a Gaussian latent representation of the kinematics learned by a feedforward encoder. The combined latent is decoded into interpretable kinematic outputs, and the resulting loss is used for gradient-based optimization. (\textbf{b}) The learned kinematics latent is replaced with randomly sampled Gaussian noise $\epsilon$, which is combined with the MT embedding and decoded to generate kinematics.}}  %MDPI: There are two subfigures (a), please check if the subfigure on the right should be ``(b)''. If so, please revise it. 
  % Authors' comments: figure revised so that the right subfigure says "(b)" rather than "(a)"
  \label{fig:kinematics_pipeline} 
  % \vspace{-3mm}
\end{figure}

To realize this, each conditional generative model was augmented with a transformer encoder layer followed by a fully connected projection layer, which encoded the conditional MT data into a representation compatible with the model’s internal embedding space (see Figure \ref{fig:kinematics_pipeline}). The resulting MT embeddings were summed into the internal embedding space, and the model was trained using a supervised loss computed against ground-truth joint-angle trajectories obtained from motion capture. This extension produced another six additional features, one for each predicted joint-angle, increasing the total feature dimensionality from 12 to 18. Hyperparameters between the MT and kinematics generators were shared.

\subsection{Methods for Classification}
Given our augmented dataset, now consisting of synthetic MT and expanded features, our goal is to distinguish among six low back movements using MT sensor data: standing extension, forward flexion, left/right lateral bending, and seated left/right rotation. No prior baselines exist for this specific sensor and task. We implemented and compared three standard architectures commonly used in related works for classifying time-series data: XGBoost \cite{Chen2016XGBoostAS}, Transformer \cite{Vaswani2017AttentionIA}, and CNN-LSTM \cite{Sainath2015ConvolutionalLS}. Comparing to IMU- or EMG-based methods would be misleading, as the sensor modalities are fundamentally different.
\begin{itemize}[leftmargin=*]
        \item For the XGBoost classifier, a tree-based model was employed, with grid search conducted over the learning rate and maximum tree depth. The optimal configuration was found to be a learning rate of 0.01 and a maximum depth of 3.
        \item For the Transformer-based classifier, grid search was performed to tune the number of encoder layers and the dimensionality of the latent space. We considered encoder depths ranging from 2 to 5 layers and latent dimensions of 6, 12, or 24. The best-performing configuration included 4 encoder layers and a latent dimension of 12.
        \item For the CNN-LSTM model, we applied grid search to optimize the hidden state dimensionality (32, 64, or 100) and the number of LSTM layers (1, 2, or 3). The optimal configuration was found to include a hidden state size of 100 and 2 LSTM layers.
\end{itemize}
\vspace{3mm}

We first evaluated these baseline models using both cross-sample test splits and leave-one-subject-out (LOSO) validation. As shown in Table~\ref{tab:splits}, we found no statistically significant difference in model performance between cross-sample and LOSO evaluations across five trials {for all three base classifiers: CNN-LSTM, Transformer, XGBoost. If substantial subject-specific leakage existed, we would expect cross-sample validation to significantly outperform LOSO due to the model exploiting subject identity markers. Furthermore, Table~\ref{tab:precission-recall} shows the precision and recall for each of the 6 exercise classes for the CNN-LSTM classifier: there is no statistical difference between the two splits for all exercises. Classifying lateral bends and flexion results in the highest precision, whereas rotation and extension have lower precision, with left rotation as the exercise most difficult to classify.} 

\begin{table}[H]
  % \setlength{\tabcolsep}{10pt}
%  \captionsetup{justification=centering}
%  \centering 
  \caption{Base classifier accuracies with different types of split {using 5-fold cross-validation.}}
  \begin{tabularx}{\textwidth}{LLLL}
  \toprule
     & \textbf{CNN-LSTM} & \textbf{Transformer} & \textbf{XGBoost} \\
    \midrule
    Subject Split & $0.899 \pm 0.027$ & $0.833 \pm 0.021$ & $0.714 \pm 0.052$ \\ 
    Sample Split & $0.917 \pm 0.038$ & $0.828 \pm 0.030$ & $0.756 \pm 0.051$ \\
   \bottomrule
  \end{tabularx}
  \label{tab:splits}
\end{table}
\vspace{-8pt}
\begin{table}[H]
%  \setlength{\tabcolsep}{7pt}
%  \captionsetup{justification=centering}
%  \centering 
  \caption{{Precision and recall per class for the base CNN-LSTM classifier with different types of split.}}
   \begin{tabularx}{\textwidth}{lLLLL}
    \toprule
  \multirow{2.5}{*}{\textbf{Exercise}}   & \multicolumn{2}{c}{\textbf{Precision}} & \multicolumn{2}{c}{\textbf{Recall}} \\   %MDPI: We merged the table rows into one cell, please confirm the suggested change.
      \cmidrule(lr){2-5} % \cmidrule(lr){4-5}
    & \textbf{Sample-Split} & \textbf{Subject-Split} & \textbf{Sample-Split} & \textbf{Subject-Split}\\  
    %MDPI: Please check whether the broken lines with gaps between two columns in the table can be modified to normal continuous table lines.
    % Authors' comments: revised to have continuous table line, rather than a gap separating the precision and recall columns
    \midrule 
    Extension & $90.1 \pm 2.5$ & $89.6 \pm 2.8$ & $88.9 \pm 3.2$ & $89.4 \pm 3.5$\\
    Flexion & $92.4 \pm 2.1$ & $91.8 \pm 2.4$ & $90.7 \pm 2.6$ & $91.3 \pm 2.2$\\
    Left Lateral Bend & $99.5 \pm 1.3$ & $98.9 \pm 1.8$ & $99.6 \pm 1.0$ & $99.8 \pm 0.9$\\
    Right Lateral Bend & $97.8 \pm 2.9$ & $99.5 \pm 2.4$ & $95.6 \pm 3.1$ & $97.2 \pm 2.7$\\
    Left Rotation & $78.6 \pm 6.9$ & $76.9 \pm 7.4$ & $82.1 \pm 5.8$ & $80.4 \pm 6.1$\\
    Right Rotation & $90.7 \pm 2.8$ & $90.1 \pm 3.0$ & $91.4 \pm 2.9$ & $90.9 \pm 2.6$\\
   \midrule
   Average & $91.5 \pm 3.1$ & $91.1 \pm 3.3$ & $91.3 \pm 3.1$ & $91.5 \pm 3.0$\\ %MDPI: Please add an explanation for the use of bold in the table footer. If the bold is unnecessary, please remove it. The following highlights in Tables 3 and 4 are the same.
   % Authors' comments: removed bold from table footer
    \bottomrule
    \end{tabularx}
    \label{tab:precission-recall}
\end{table}

\begin{figure}[H]
%  \captionsetup{justification=centering}
%  \centering 
  \includegraphics[width=0.6\textwidth]{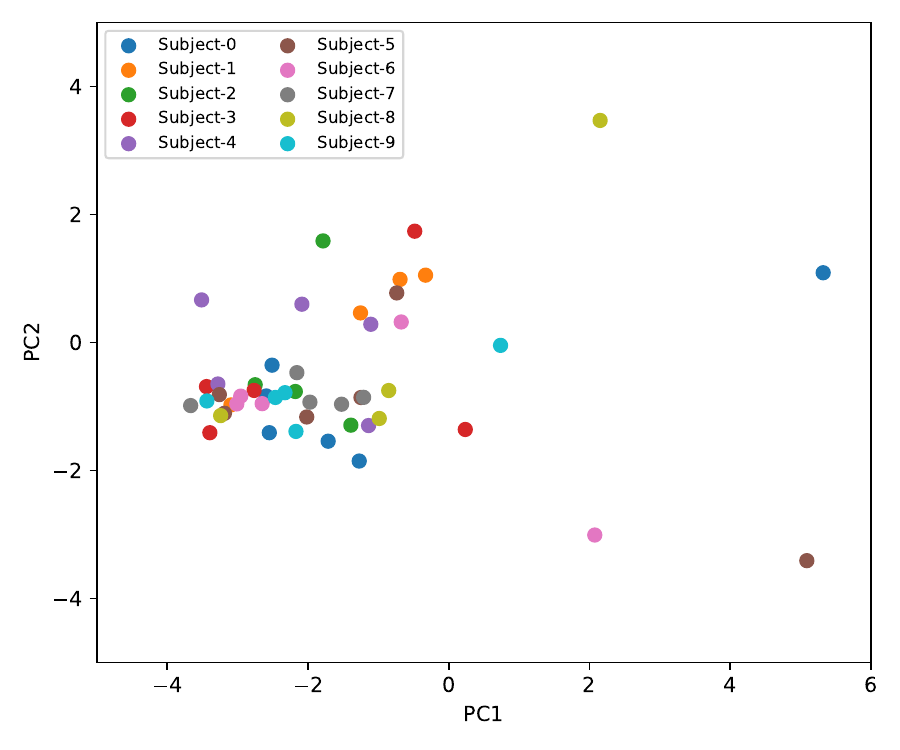} 
  \caption{{PCA analysis of samples colored by subject.}}
  \label{fig:pca} 
  % \vspace{-3mm}
\end{figure}

{To further investigate if there is subject-specific information in our dataset, we trained a dedicated subject classifier using 5-fold cross-validation, achieving only 11\% ± 5.2 accuracy (the chance level is 10\% for 10 subjects). This near-random performance confirms that subject-specific signatures are not strongly encoded in the features used for our primary classification task. PCA is also useful in capturing data variance---the PCA visualization from Figure~\ref{fig:pca} demonstrates that subject clusters are not well-separated in this reduced feature space. Individual subjects' samples are distributed throughout the feature space rather than forming distinct, isolated clusters. This indicates that subject variance is not a dominant source of variation in our features. Even sample outliers are not specific to one subject but rather appear scattered across multiple subjects. Our preprocessing pipeline was designed to extract exercise-related patterns while suppressing subject-specific confounds: Hampel filtering using local windows within each sample removes outlier artifacts and eliminates subject-specific noise patterns, and per movement min-max normalization removes subject-specific baseline offsets and amplitude identification markers.}

{As there is no significant difference between cross-sample and LOSO evaluations in our dataset,} we adopted cross-sample splits for subsequent experiments, providing greater flexibility in train and test dataset composition and yielding clearer insights into the data requirements for model learning. We report mean ± std over five trials in each table to demonstrate consistency. Given our dataset of 180 samples and 10 subjects, a subject-split leaves 18 samples per subject, which limits flexibility for experimenting with different fold sizes. The use of stratified sampling allows to better understand how much data the model requires to reach optimal performance. LOSO evaluation will be important in subsequent studies on a larger population, including heterogeneous LBP characteristics.

\section{Results and Discussion}

To evaluate our pipeline, we must look differently at the conditional generative models and classification models used within. For generative models, we evaluate on typical time-series distance functions: Wasserstein Distance and Frechet Time-Series Distance (FTSD)~\cite{driemel2015clustering}. For classification models, we evaluate based on accuracy.

To summarize from the previous sections, our dataset consists of 10 participants, \linebreak  6 movement types, and 3 repetitions per movement type, resulting in 180 original MT labeled samples. The labels are the specific movements supervised by a physical therapist in the lab. For data augmentation, we additionally generate 20 samples per movement type, resulting in 120 synthetic MT samples, thus bringing the total to 300 MT samples. The original features consist of 6 MT time series, corresponding to the 6 sensors on the low back. For feature augmentation, DTFT produces 6 additional features, and the generated kinematics produce another 6 features, bringing the total to 18 features. 

\subsection{Generated Motion Tape Evaluation}
To evaluate our synthetic MT time-series data, we assess both distributional differences compared to real MT data and overall temporal alignment using Wasserstein and FTSD metrics. Evaluating on both metrics allows for richer insights into the structure and similarity of the sequences.

The Wasserstein Distance, also known as Earth Movers Distance, compares the shapes of two time-series distributions $A$ and $B$, even if they do not align in time. It is commonly referred to as measuring the minimum ``cost'' of transforming one distribution into \mbox{the other:} %MDPI: Please make sure that: - There are no duplicated equations; - Variables are consistently formatted (for example, italic in all places in the paper).
% Authors' comments: checked

\begin{linenomath}
\begin{equation}
W(A, B) = \inf_{\pi\in\Gamma(A,B)} \int_{R\times R} |x - y|d\pi(x, y),
\end{equation}
\end{linenomath}
where $\Gamma(A, B)$ is the set of probability spaces whose marginal distributions correspond to $A$ and $B$ respectively.

To evaluate our synthetic data considering temporal structure, we choose FTSD. This metric measures the similarity between distributions by taking into account the location and ordering of points at certain steps in time:
\begin{linenomath}
\begin{equation}
F(A, B) = \inf_{\alpha,\beta} \left\{ \max_{t\in[0,1]} \{d(A(\alpha(t)), B(\beta(t)))\} \right\},
\end{equation}
\end{linenomath}
where $d$ is the distance function between two points in the metric space of curves $A$ and $B$, and $\alpha$ and $\beta$ are all reparameterizations of $A$ and $B$ respectively.

To evaluate our generative models, we used all MT data for training, generating \mbox{120 new} samples after convergence. This was repeated 5 times per model, with randomly initialized weights, to calculate average and standard deviation metrics for Wasserstein Distance and FTSD. As seen in Table~\ref{tab:2}, the C-VAE and Diffusion-TS models significantly outperform the TimeGAN model for generating synthetic MT data.

\begin{table}[H]
%  \setlength{\tabcolsep}{7pt}
  % \centering 
  \caption{{Comparison of MT and kinematics generative models.} The FTSD and Wasserstein distance metrics measure the similarity between predicted and ground-truth time series (generated MT vs. real MT and generated kinematics vs. real MoCap kinematics). The bold values denote the best model for each scenario. C-VAE and Diffusion-TS models have similar performances, but both significantly outperform TimeGAN models.}
  \begin{tabularx}{\textwidth}{LLLLL}
  \toprule
 \multirow{2}{*}{\textbf{Model}}    & \multicolumn{2}{c}{\textbf{Motion Tape}} & \multicolumn{2}{c}{\textbf{Kinematics}} \\ %MDPI: We merged the table rows into one cell, please confirm the suggested change.
    \cmidrule(lr){2-5} % \cmidrule(lr){4-5}
     & \textbf{FTSD} & \textbf{Wasserstein} & \textbf{FTSD} & \textbf{Wasserstein} \\ %MDPI: Please check whether the broken lines with gaps between two columns in the table can be modified to normal continuous table lines.
     % Authors' comments: revised to be one continuous line
    \midrule
    C-VAE & $35.953 \pm 9.744$ & $2.498 \pm 1.210$ & \textbf{9.560} $\pm$ \textbf{4.022} & \textbf{0.275} $\pm$ \textbf{0.097} \\ 
    Diffusion-TS & \textbf{19.418} $\pm$ \textbf{6.727} & \textbf{1.040} $\pm$ \textbf{0.406} & $13.565 \pm 5.422$ & $0.621 \pm 0.261$ \\
    TimeGAN & $60.717 \pm 16.110$ & $6.822 \pm 2.982$ & $32.211 \pm 7.002$ & $1.628 \pm 0.742$ \\
    % Authors' comments: added explanation about bold in the figure caption.
   \bottomrule
  \end{tabularx}
  \label{tab:2} 
\end{table}

\subsection{Generated Kinematics Evaluation}
\begin{figure}[H]
  % \centering
  \includegraphics[width=\textwidth]{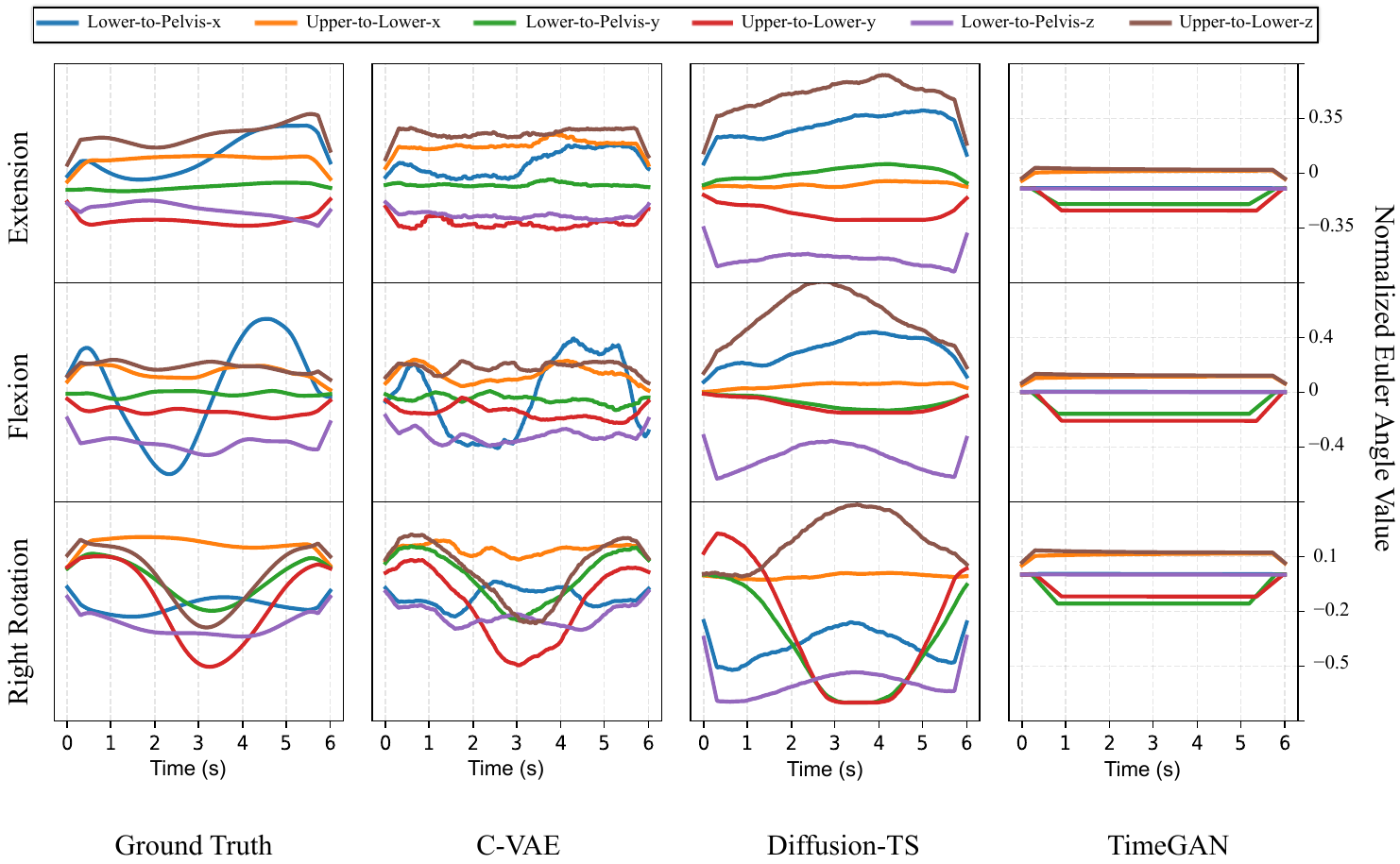}
  \caption{{Examples of generated kinematics compared with ground truth.} Each row compares different low back movements and each column compares different model kinematic predictions for that row's movement. The final column displays the mode collapse exhibited by TimeGAN.}
  \label{fig:sensors_kine_vs_gt}
  % \vspace{-3mm}
\end{figure}

We evaluate the generated kinematic time-series data using the same metrics as our generated MT data. Since the kinematic joint angles are a function of the MT data, it is expected that the distance to ground truth in generated kinematics samples would be smaller compared to our MT evaluations. Table~\ref{tab:2} and Figure~\ref{fig:sensors_kine_vs_gt} confirm this expectation and also show the superiority of C-VAE and Diffusion-TS for the kinematic translation task. On the other hand, TimeGAN exhibited mode collapse, generating highly similar kinematic outputs despite the variation of input noise or movement.
\vspace{-5pt}

\subsection{Classification Evaluation}

To assess the performance of our classification models, we evaluate them using accuracy. Given that our dataset is balanced and the primary objective is to effectively classify lower back movements, accuracy provides the most informative measure of model performance. When evaluating our models, we randomly select stratified MT samples making up 25\% of all real data to be part of a test set. The rest of the real data along with synthetic data (depending on the experiment) is used for training. This procedure is conducted \mbox{5 times} per model, with random weight initialization, where an average accuracy score and standard deviation is recorded.

\begin{figure}[H]
  % \centering 
  \includegraphics[width=0.99\textwidth]{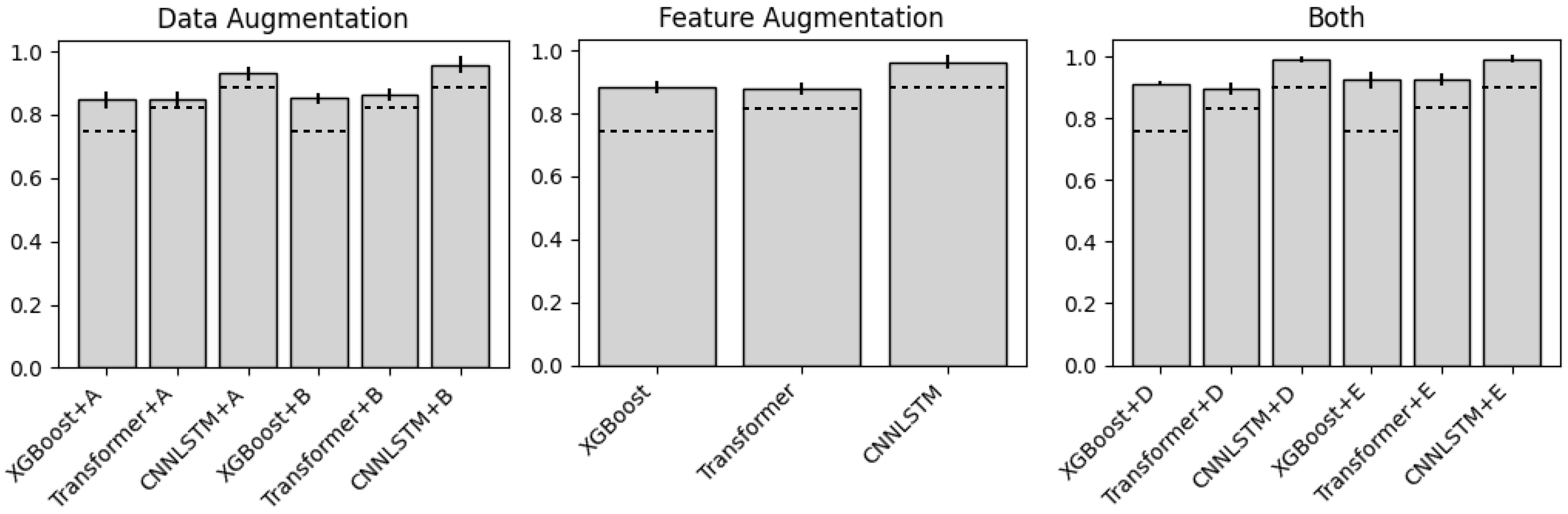} 
  \caption{{Accuracy barplots evaluating the effect of different augmentations.} Models with \texttt{+A}/\texttt{+B} use C-VAE/Diffusion-TS to generate synthetic MT data. Models with \texttt{+D}/\texttt{+E} use C-VAE/Diffusion-TS to generate synthetic MT data and the same model to generate conditional kinematics data. The dotted line on each plot represents that classifier's base classification accuracy (no augmentations).}
  \label{fig:5} 
  % \vspace{-3mm}
\end{figure}

In Table~\ref{tab:3} and Figure~\ref{fig:5} we compare the accuracy scores of different combinations of synthetic and classification models. We exclude TimeGAN as synthetic MT and kinematic translation modules due to poor performance recorded in Table~\ref{tab:2}. As this is the first ML study for low back movement classification with MT data, no prior baselines exist for this specific task and sensor. We use true kinematics as an ablation study to assess model performance with ideal inputs, differentiating classification capability from estimation noise: feature augmentation in Table~\ref{tab:3} represents augmentation with true kinematics and DTFT features, whereas MT-AIM uses data augmentation and generated kinematics.  The reason for using real kinematics for Feature Augment is to show the classification performance upper bound using only feature augmentation. It validates that kinematic information is valuable and supports our design choice to generate kinematics, showing the maximum possible benefit with just feature augmentation if generation were perfect. 
%During classification, the models do not have access to the test kinematics data. 

\begin{table}[H]
  % \setlength{\tabcolsep}{10pt}
  % \centering 
  \caption{{Classification accuracies of different models for low back movements.} Data augmentation is the base classifier with the addition of synthetic MT data. Feature augmentation is the base classifier with the addition of real (rather than predicted) kinematics and DTFT feature augmentations. MT-AIM uses predicted MT and predicted kinematics from our best generators. Only the highest-performing combinations of data and feature augmentation models are shown (TimeGAN is left out). The bold values denote the best model for each scenario.}
  \begin{tabularx}{\textwidth}{lllll}
  \toprule
    \textbf{Model} & \textbf{Base} & \textbf{Data Augment} & \textbf{Feature Augment} & \textbf{Both (MT-AIM)} \\
    \midrule
    XGBoost & $0.756 \pm 0.051$ & $0.852 \pm 0.017$ & $0.883 \pm 0.018$ & $0.926 \pm 0.024$ \\ 
    Transformer & $0.828 \pm 0.030$ & $0.864 \pm 0.019$ & $0.879 \pm 0.020$ & $0.929 \pm 0.020$ \\
    CNN-LSTM & \textbf{0.917} $\pm$ \textbf{0.038} & \textbf{0.957} $\pm$ \textbf{0.027} & \textbf{0.964} $\pm$ \textbf{0.020} & \textbf{0.994} $\pm$ \textbf{0.011} \\
        % Authors' comments: added explanation about bold in the figure caption.
   \bottomrule
  \end{tabularx}
  \label{tab:3} 
\end{table}
\vspace{-12pt}

From the visualizations, we can see that both augmenting the data and features significantly impacts the performance of our classification models. Furthermore, the models using the CNN-LSTM classification both reached over 99\% average accuracy over the five trials. We also see from Figure~\ref{fig:5} that choosing Diffusion-TS over C-VAE, or vice versa, did not significantly impact performance; rather, the existence of augmentation is all that mattered. In general, more metrics beyond accuracy may provide a more comprehensive evaluation, as class-wise performance may vary. However, in this study we managed to train a model to near perfect accuracy; thus, our results highlight the validity of MT and augmentation approaches.  
\section{Limitations and Opportunities for Future Work}
% Explain when your approach may not apply, or things you could not
% check.  \emph{Discussing limitations is essential.  Both ACs and
%   reviewers have been advised to be skeptical of any work that does
%   not consider limitations.}

%Although MT-AIM is an effective pipeline for classifying MT data, we highlight the following limitations and propose suggestions for future research.

MT-AIM offers an effective pipeline for classifying lower back movements using MT. The standard approach for biometric monitoring technologies is to first conduct a study on a healthy population and then validate it in a clinical population. This study realizes the first step in a healthy population so that we can ensure full ranges of movement in all movement directions. These data were derived from a preliminary laboratory validation study under the supervision of a physical therapist; the movement types for this study were selected by a physical therapist as clinically relevant for LBP assessment. Future clinical validation studies are needed in larger samples and in people with and without LBP. Future work includes comparing the movements between people with and without LBP, identifying impairments and subtypes of LBP based on MT, and testing functional activities in free-living environments. Our work is promising in providing a framework for classification, with the goal to enable doctors to remotely assess and recommend physical therapy for LBP. 

While many aspects need to be solved before this approach can be used in free-living environments for patient monitoring to inform decision making and improve outcomes, this classification pipeline is an important step in that direction. The clinical problem of managing LBP requires: evaluating movements to determine how movement is associated with the LBP problem, estimating the strain on the lower back throughout the day, assessing whether patients are performing exercises correctly, and assessing progression throughout treatment and risk of LBP development or recurrence. A new generation of MT is under development that transmits data wirelessly to a smartphone app, which sends it to the cloud for real-time monitoring. Furthermore, in our experiments, inference speed was quicker than 20 ms which is faster than the frequency of the sensor readings, so the computational cost is negligible for the task of classifying movements. However, experiments involving real-time monitoring and for longer duration will need new models to analyze compound movements with MT and assess progression throughout treatment, whose computational cost will need to be evaluated.

The small sample size limits our ability to conclude the model generalization to a population of people performing lower back movements. External datasets do not exist for MT, as it is a novel sensor. Small sample size is common at this stage of development for new sensors, as the sample size is purposely small because the data gathered is used to help improve sensors for future larger studies. Different people could have different MT readings based on height, weight, and other demographics, so more subjects are needed with a variety of body types. For example, individuals with obesity may have different skin strains, leading to difficulty in interpreting their MT readings and classifying the movements. Furthermore, individuals with LBP may have different movement patterns. Also, the amount of noise and size of dataset remained constant in our research, so we can't make any conclusions about MT-AIM's performance on other datasets of varying size \linebreak  and fidelity.

Variability in the GNS coverage on different MT sensors may cause inconsistent sensitivity, which can lead to data samples that are hard to classify.
We are exploring new materials for fabricating MT using Multiwalled Carbon Nanotubes, which are promising in being less noisy than the GNS used in this study. However, there is a tradeoff between sensitivity and signal-to-noise ratio; thus, some noise is to be expected. The advantage of our approach is that even when the sensing technology is noisy, the framework still reaches very high accuracy. 

Furthermore, MT baseline resistance values can drift over time. This can cause classification issues if the movement profiles drift. Since measurements were taken in a controlled setting in the lab, sensor drift was not a concern in our study. Each measurement session lasted about an hour per participant, including lab setup and interviews. We also normalized the sensors to baseline before each movement type. There is promising research to mitigate the sensor drift in the classification pipeline during prolonged use in \mbox{free-living environments}.

{Because MT operates in response to local skin strain, variations in sensor placement can influence both signal magnitude and temporal characteristics. Changes in placement relative to underlying muscle groups may alter the local strain field experienced by the sensor, leading to differences in sensitivity to specific movements. This makes placement critical, as only the targeted muscle group or bony segment should be covered by the MT sensing region. In the event of misplacement, undesirable perturbations may arise from adjacent or unintended muscle groups within the same region. Similarly, orientation affects how skin deformation is transferred into the conductive network, which can impact repeatability across sessions or subjects.}

{It could be challenging for non-experts to reliably identify anatomical landmarks, but clinicians are expertly trained to do this. Therefore, in this study the sensors were placed on all participants by a physical therapist. Future studies that need non-expert application could create a template that is based on easily identified landmarks---like the PSIS (Posterior Superior Iliac Spine) visible as dimples on the back of the pelvis---and then the MTs are laid down according to the template.}

\textls[-15]{The task of generating kinematic joint angles from MT data still has a lot of room for improvement. Physics-informed training seems like a promising avenue. This could lead to 3D modeling of human motion using MT. Another suggestion for future work would be to use MT for form analysis in athletic performance such as weight lifting. It would be interesting to see if the tape is precise enough to recognize small anomalies in human motion.}

%\begin{itemize}
%  \item The task of generating kinematic joint angles from MT data still has a lot of room for improvement. Physics-informed training seems like a promising avenue. This could lead to 3D modeling of human motion using MT.
%  \item Small sample size limits our ability to conclude any model generalization to a population of people performing lower back movements. Different people could have drastically different MT readings based on height, weight, and other demographics, so more subjects are needed.
%  \item Amount of noise and size of dataset remained constant in our research, so we can't make any conclusions about MT-AIM's performance on other datasets of varying size and fidelity.
%  \item Variability in the GNS coverage on different MT may cause inconsistent sensitivity, which can lead to data samples that are hard to classify. Furthermore, MT baseline resistance values over time can drift. This can cause classification issues if the movement profiles drift; however, normalization of sensors to baseline for each movement currently mitigates sensor drift. There is promising research to mitigate the sensor drift in the classification pipeline.
%  \item Another suggestion for future work would be to use MT for form analysis in athletic performance such as weight lifting. It would be interesting to see if the tape is precise enough to recognize small anomalies in human motion.
%\end{itemize}

\section{Conclusions}
We developed MT-AIM, a deep learning pipeline to classify low back movement based on MT sensors. MT-AIM demonstrates the potential of using low-cost, portable wearable sensors, like MT, for classifying human motion, while also providing a viable solution for remote monitoring of lower back strain in people with LBP. By leveraging synthetic data generation and feature augmentation, MT-AIM addresses the challenges posed by the small scale and noise of MT data, achieving state-of-the-art performance despite \linebreak  these limitations.

The ability to generate synthetic MT data of a desired movement, alongside predictions of joint kinematics, significantly enhances the accuracy of our classification models. Moreover, this approach may not only aid in the remote monitoring of low back strain but also open the door to future applications with other wearable sensors.

However, the limited dataset suggests that further validation with larger, more diverse datasets is necessary. As this technology evolves, extending the model to incorporate more nuanced features, such as physics-based regularization terms, could lead to more complex insights on human motion modeling with MT. Future work should focus on validating the system with patients experiencing LBP and exploring real-world deployment scenarios to assess long-term reliability and clinical utility.

%%%%%%%%%%%%%%%%%%%%%%%%%%%%%%%%%%%%%%%%%%
\vspace{6pt} 
\authorcontributions{Conceptualization, R.Y., E.F., S.P.G. and K.J.L; Methodology, J.L., R.Y., E.F., and A.L.; Software, J.L., E.W. and S.P.G.; Validation, S.P.G.; Formal analysis, J.L., R.Y. and E.F.; Investigation, J.L., R.Y., E.F., A.L., S.P.G., E.W., and K.J.L.; Resources, E.F., S.P.G., K.J.L. and R.Y.; Data curation, E.W. and S.P.G.; Writing---original draft, J.L. and R.Y.; Writing---review \& editing, J.L., E.F., A.L., E.W., S.P.G, and K.J.L.; Visualization, J.L. and E.W.; Project administration, E.F. and S.P.G.; Funding acquisition, E.F., S.P.G., K.J.L., and R.Y.. All authors have read and agreed to the published version of the manuscript.} %MDPI: We added the Author Contributions based on the information submitted online at susy.mdpi.com, and we changed the authors’ names into an abbreviated format. Please confirm.
% Authors' comments: revised.

\funding{This research was supported by the U.S. National Science Foundation (NSF) under grant no. IIS-2205093 (PI: E.F.).} %MDPI: We seperate the Acknowledgments as Funding, IRBS and ICS. Please confirm the suggested change. If there is any other acknowledgments, please revise it.
% Authors' comments: confirmed.

\institutionalreview{The study was conducted in accordance with the Declaration of Helsinki, and the protocol was approved by the San Diego State University Institutional Review Board (Protocol Number: HS-2022-0269; Approval Date: 25 January 2023).}   

\informedconsent{\textls[-15]{Informed consent was obtained from all subjects involved in the study.}}

%%%%%%%%%%%%%%%%%%%%%%%%%%%%%%%%%%%%%%%%%%
\dataavailability{The de-identified dataset used in this study is available upon appropriate request to the corresponding author. The data are not publicly available, owing to ethical concerns. The complete source code for MT-AIM is available in a public GitHub repository \cite{levy2025msadapt}.}
% will be made available in a public GitHub repository
% start a new page without indent 4.6cm
%\clearpage

%\acknowledgments{  } 

\conflictsofinterest{Co-author Kenneth J. Loh is a co-founder of JAK Labs, Inc., a company that may potentially benefit from the research results. JAK Labs intends to commercialize Motion Tape for different physical exercise and sports markets, among others. The terms of this arrangement have been reviewed and approved by the University of California San Diego in accordance with its conflict-of-interest policies.} 

%%%%%%%%%%%%%%%%%%%%%%%%%%%%%%%%%%%%%%%%%%
\abbreviations{Abbreviations}{
The following abbreviations are used in this manuscript:

\vspace{6pt}
\noindent 
\begin{tabular}{@{}ll}
    AI & Artificial Intelligence\\
    CNN-LSTM & Convolutional Neural Network Long Short-Term Memory\\
    C-VAE & Conditional Variational Autoencoder\\
    DTFT & Discrete-Time Fourier Transform\\
    EMG & Electromyography\\
    FTSD & Frechet Time-Series Distance\\
    GAN & Generative Adversarial Network\\
    GNS & Graphene nanosheet\\
    K-tape & Kinesiology tape\\ 
    ICA & Independent Component Analysis\\
    IMU & Inertial Measurement Unit\\
    LBP & Low back pain\\
    LOSO & Leave-One-Subject-Out\\
    ML & Machine Learning\\
    MoCap & Motion Capture\\
    MT & Motion Tape\\
    MT-AIM & Motion-Tape Augmentation Inference Model\\
    PCA & Principal Component Analysis
    \end{tabular}
}

\begin{adjustwidth}{-\extralength}{0cm}

%References have been checked and layouted accordingly. Please, go over the entire manuscript and check whether all of the references are cited and listed correctly and in order.  
%   
%Please, DO NOT change reference layout. May you need to do any changes in the references, please, write a comment as the reference list has undergone thorough layout editing.  
%   
%Please NEVER use EndNote or other tools to re-arrange the reference order.  
\reftitle{References}

\PublishersNote{}
\end{adjustwidth}

\end{document}